\documentclass[conference]{IEEEtran}
\IEEEoverridecommandlockouts
\usepackage{cite}
\usepackage{amsmath,amssymb,amsfonts}
\usepackage{algorithmic}
\usepackage{graphicx}
\usepackage{textcomp}
\usepackage[square,sort,comma,numbers]{natbib}
\usepackage{xcolor}

\usepackage{booktabs}
\usepackage{graphicx} 
\def\BibTeX{{\rm B\kern-.05em{\sc i\kern-.025em b}\kern-.08em
    T\kern-.1667em\lower.7ex\hbox{E}\kern-.125emX}}
\begin{document}

\title{Multiple Run Ensemble Learning with Low-Dimensional Knowledge Graph Embeddings
}
\author{\IEEEauthorblockN{Chengjin Xu\IEEEauthorrefmark{1}, Mojtaba Nayyeri\IEEEauthorrefmark{1}, Sahar Vahdati\IEEEauthorrefmark{2} and Jens Lehmann\IEEEauthorrefmark{1}\IEEEauthorrefmark{3}}
\IEEEauthorblockA{\IEEEauthorrefmark{1}SDA Research, University of Bonn, Bonn, Germany\\
Email: \{xuc, nayyeri, jens.lehmann\}@cs.uni-bonn.de}
\IEEEauthorblockA{\IEEEauthorrefmark{2}University of Oxford, Oxford, UK\\
Email: sahar.vahdati@cs.ox.ac.uk}
\IEEEauthorblockA{\IEEEauthorrefmark{3}Fraunhofer IAIS, Bonn, Germany
\\
Email: jens.lehmann@iais.fraunhofer.de}
}


\maketitle

\begin{abstract}
Knowledge graphs (KGs) represent world facts in a structured form. Although knowledge graphs are quantitatively huge and consist of millions of triples, the coverage is still only a small fraction of world’s knowledge. Among the top approaches of recent years, link prediction using knowledge graph embedding (KGE) models has gained significant attention for knowledge graph completion. Various embedding models have been proposed so far, among which, some recent KGE models obtain state-of-the-art performance on link prediction tasks by using embeddings with a high dimension (e.g. 1000) which accelerate the costs of training and evaluation considering the large scale of KGs. In this paper, we propose a simple but effective performance boosting strategy for KGE models by using multiple low dimensions in different repetition rounds of the same model. For example, instead of training a model one time with a large embedding size of 1200, we repeat the training of the model 6 times in parallel with an embedding size of 200 and then combine the 6 separate models for testing while the overall numbers of adjustable parameters are same (6*200=1200) and the total memory footprint remains the same. We show that our approach enables different models to better cope with their expressiveness issues on modeling various graph patterns such as symmetric, 1-n, n-1 and n-n. In order to justify our findings, we conduct experiments on various KGE models. Experimental results on standard benchmark datasets, namely FB15K, FB15K-237 and WN18RR, show that multiple low-dimensional models of the same kind outperform the corresponding single high-dimensional models on link prediction in a certain range and have advantages in training efficiency by using parallel training while the overall numbers of adjustable parameters are same. 
\end{abstract}


\section{Introduction}
Numerous knowledge graphs including lexical datasets and world's knowledge, such as WordNet \cite{b1}, FreeBase~\cite{b2}, YAGO~\cite{b3}, and DBpedia \cite{b4}, have been published with different utilization purposes. These KGs have become a significant resource for many AI-based applications such as question answering and recommendation systems~\cite{b5}. 
Therefore, a new horizon for using machine learning approaches on structured data at scale has been opened up for leading science and industry.

Despite all the advantages of KGs in down stream tasks, one of the main challenges of existing KGs is their incompleteness~\cite{b6}. 
Knowledge graph completion using link prediction approaches aims at addressing the incompleteness of KGs. 
Among various link prediction approaches, knowledge graph embedding (KGE) has gained significant attention recently. 
A KGE model takes a KG in the form of triple facts $(h,r,t)$ where $h,t$ are entities (nodes) and $r$ is a relation (link) between the entities (e.g.~(\textit{Berlin, Capital\_of, Germany})). 
A vector is assigned to each element of a triple ($\mathbf{h}, \mathbf{r}, \mathbf{t}$) in a KG and all vectors are then adjusted by optimizing a loss function. The likelihood of a triple is then measured by using a score function over the embedding vectors ($\mathbf{h}, \mathbf{r}, \mathbf{t}$). To evaluate the performances of KGE models on link prediction, several benchmarks including FB15K, WN18~\cite{b6}, FB15K237 and WN18RR~\cite{b7} are established by extracting subsets from large-scale KGs, e.g., WordNet~\cite{b8} and FreeBase~\cite{b9}.

The score functions of models play an important role in the performance of KGE models. Numerous KGE models with a focus on score functions have been proposed.
Among them, several recent models, including ComplExN3~\cite{b10}, RotatE~\cite{b11} and QuatE~\cite{b12} take the advantage of hyper-complex representations and high-dimensional embeddings to achieve the state-of-the-art results. The high performances of RotatE, ComplExN3 and QuatE on FB15K are achieved by using embeddings with the dimensions of 1000, 1000 and 2000. Due to the usage of complex-valued vectors and quaternion vectors of ComplExN3 and QuatE, their best performing settings result in 4000 adjustable parameters for each entity and relation. On the other hand, for KGE models with higher embedding dimensions, there is a risk of the redundancy of parameters leading to unnecessary consumption on training time as well as memory space while the generalisation capabilities of the learned models are not improved. For example, the experimental results of ComplExN3 shows that the performances of a ComplExN3 with 500-dimensional embeddings on FB15K237 and WN18RR are almost same as a ComplExN3 with a embedding dimension of 2000. 

Such observations show that the above mentioned state-of-the-art models take advantages of high embedding dimensions and multiple-vector representation resulting in a large embedding size $d$ which is equal to the total number of adjustable parameters of each entity/relation embedding. 
In other words, these models use \textbf{single} model with a multi-part \textbf{high dimensional} embeddings (the total embedding size is $1 \times d_h$). In contrast to those models, we use the same model \textbf{multiple} ($k$) times in parallel trainings with \textbf{low dimension} (the overall embedding size is $k \times d_l$). In order to have a fair comparison, we enforce $k \times d_l = 1\times d_h$ to guarantee that the overall numbers of adjustable parameters of multiple low-dimensional models is equal to the single high-dimensional model.
Using multiple low dimensional model instead of a single high dimensional model improves the expresivity of various models in handling various patterns including symmetric, 1-n, n-1 and n-n. The experimental results on three benchmarks show that the ensemble of the same KGE model several times trained with low-dimensions results in a better performances than training that model once with a high dimension, regarding link prediction accuracy (higher expectations and lower standard deviations) and training time.
\section{Related Work}\label{Related Work}
KGE models can be roughly classified into two groups, distance-based models and semantic matching models~\cite{b5}. Here we review two distance-based KGE models, i.e., TransE, RotatE and three semantic matching models i.e., DistMult, ComplEx and ComplExN3.
Each KGE model defines a score function $f(h,r,t)$ which takes embedding vectors of a triple $(\mathbf{h}, \mathbf{r}, \mathbf{t})$ and returns a value showing the degree of correctness of the triple.
\\
\textbf{TransE} \cite{b6} computes the score of a triple $h,r,t$ by measuring the distance between relation-specific translated head ($\mathbf{h} + \mathbf{r}$) and the tail $\mathbf{t}$ as
\begin{align}\label{eq:translation}
     f(h,r,t) = -\| \mathbf{h} + \mathbf{r} - \mathbf{t} \|,
\end{align}
to enforce $\mathbf{h} + \mathbf{r} \approx \mathbf{t}$ for each positive triple ($h,r,t$) in the vector space. The TransE model embeds entities/relation into $d$ dimensional real space, i.e., $\mathbf{h}, \mathbf{r}, \mathbf{t} \in \mathbb{R}^{d} $. 
\\
\textbf{RotatE} \cite{b11} aims at mapping each element of the head embedding ($\mathbf{h}_i$) to the corresponding tail embedding $\mathbf{t}_i$ by using relation-specific rotation $\mathbf{r}_i = e^{i\theta}.$
The score of each triple $(h,r,t)$ is computed as
\begin{equation}
    f(h,r,t) = -\| \mathbf{h}\circ\mathbf{r} - \mathbf{t} \|,
\end{equation}
where $\circ$ is the element-wise multiplication and $\mathbf{h}, \mathbf{r}, \mathbf{t} \in \mathbb{C}^d$. This enforces $\mathbf{h} \circ \mathbf{r} \approx \mathbf{t}$ for each positive triple ($h,r,t$). 
\\
\textbf{DistMult} \cite{b13} is based on the Bilinear model~\cite{b17} where each relation is represented by a diagonal matrix rather than a full matrix. The formulation of score function is
\begin{equation}\label{eq:DistMult_score}
    f(h,r,t) = \mathbf{h}^{\top}\mathbf{diag(r)}\mathbf{t} = \sum_{i=0}^{d}  \mathbf{h}_i \cdot  \mathbf{r}_i \cdot \mathbf{t}_i.
\end{equation}
This score captures pairwise interactions between only the components of $\mathbf{h}$ and $\mathbf{t}$ along the same dimension and thus can only deal with symmetric relations.
\\
\textbf{CompEx} \cite{b14} was proposed as an elegant way to solve the shortcoming of DistMult in modeling asymmetric relation. Its main contribution is to embed KGs in complex space. The score function is defined as 
\begin{equation}
\begin{split}
f(h,r,t) = \mathrm{Re}(\mathbf{h}^{\top}\mathbf{diag(r)}\mathbf{\overline{t}}) = \mathrm{Re}\left(\sum_{i=1}^{d} \mathbf{h}_{i}\mathbf{r}_{i}\bar{\mathbf{t}}_{i}   \right),  
\label{eq:comlex}
\end{split}
\end{equation}
where $ \mathbf{r}, \mathbf{h}, \mathbf{t} \in \mathbb{C}^{d}$, $\mathbf{t}_{i}$ represents the complex conjugate of $\mathbf{t}$. By using this scoring function, triples that have asymmetric relations can obtain different scores.
\textbf{ComplExN3}~\cite{b10} extends ComplEx with weighted nuclear 3-norm (N3 regularization) and uses a multi-class logistic loss function as optimization objective to achieve the state-of-the-art results.


Krompass and Tresp~\cite{b15} integer multiple different KGE models into one score function and combines them during the training phase. Likewise, Muroemagi et al.~\cite{b16} combine different word embedding models using an iterative method. However, we focus on stretching and squeezing the dimensions of the same models which are trained separately and combined only for testing. 
\section{Proposed Approach}
Compared to TransE and DistMult, ComplExN3, RotatE and QuatE take advantages of high-dimensional multi-part embeddings to achieve state-of-the-art performances on link prediction. In these models, all parts of embeddings are adjusted simultaneously. By contrast,
in this part we propose a new approach which combines \textbf{multiple} models of the same kind where each model contains \textbf{low-dimensional} embeddings and is trained separately.

Let us have a model $\mathcal{M}$ (e.g.~RotatE) with the embedding size $d$ where $d$ is equal to the total number of adjustable parameters of each entity/relation embedding in a KGE model. Let $\mathcal{E}$ denote the set of all entities and $\mathcal{R}$ the set of
all relations present in a knowledge graph. A triple is represented as $(h, r, t)$, with $h$, $t$ $\in \mathcal{E}$ denoting head and tail entities respectively and
$r\in \mathcal{R}$ the relation between them. We use $\Omega = \{(h,r,t)\} \subseteq \mathcal{E}\times\mathcal{R}\times\mathcal{E}$ to denote the set of observed triples. We follow the steps below in our approach:



\begin{itemize}
    \item[(a)] We first generate $k$ times copies of an underlying  model $\mathcal{M}$. The $j$th copy of the model is denoted by $\mathcal{M}_j, j=1, \ldots, k,$ and the corresponding $d_l$ dimensional embeddings of $(h,r,t)$ are denoted by $(\mathbf{h}_j, \mathbf{r}_j, \mathbf{t}_j).$ The vectors are randomly initialized before training.
    \item[(b)] We then train each of the models $\mathcal{M}_j, j=1, \ldots,k$ separately with the same loss function. The training process can speed up by using parallel computing.
    \item[(c)] Finally, the testing is performed by using the following score function
    \begin{equation}
        f(h,r,t) = \frac{1}{k}\sum_{j=1}^{k} f_{\mathcal{M}}(\mathbf{h}_j, \mathbf{r}_j, \mathbf{t}_j),
    \end{equation}
    where $f_{\mathcal{M}}(\mathbf{h}_j, \mathbf{r}_j, \mathbf{t}_j)$ is the score of a triple $(h,r,t)$ computed by the $j$th copy of the model.
\end{itemize}

\begin{figure}[t!]
\centering
\includegraphics[width=8cm]{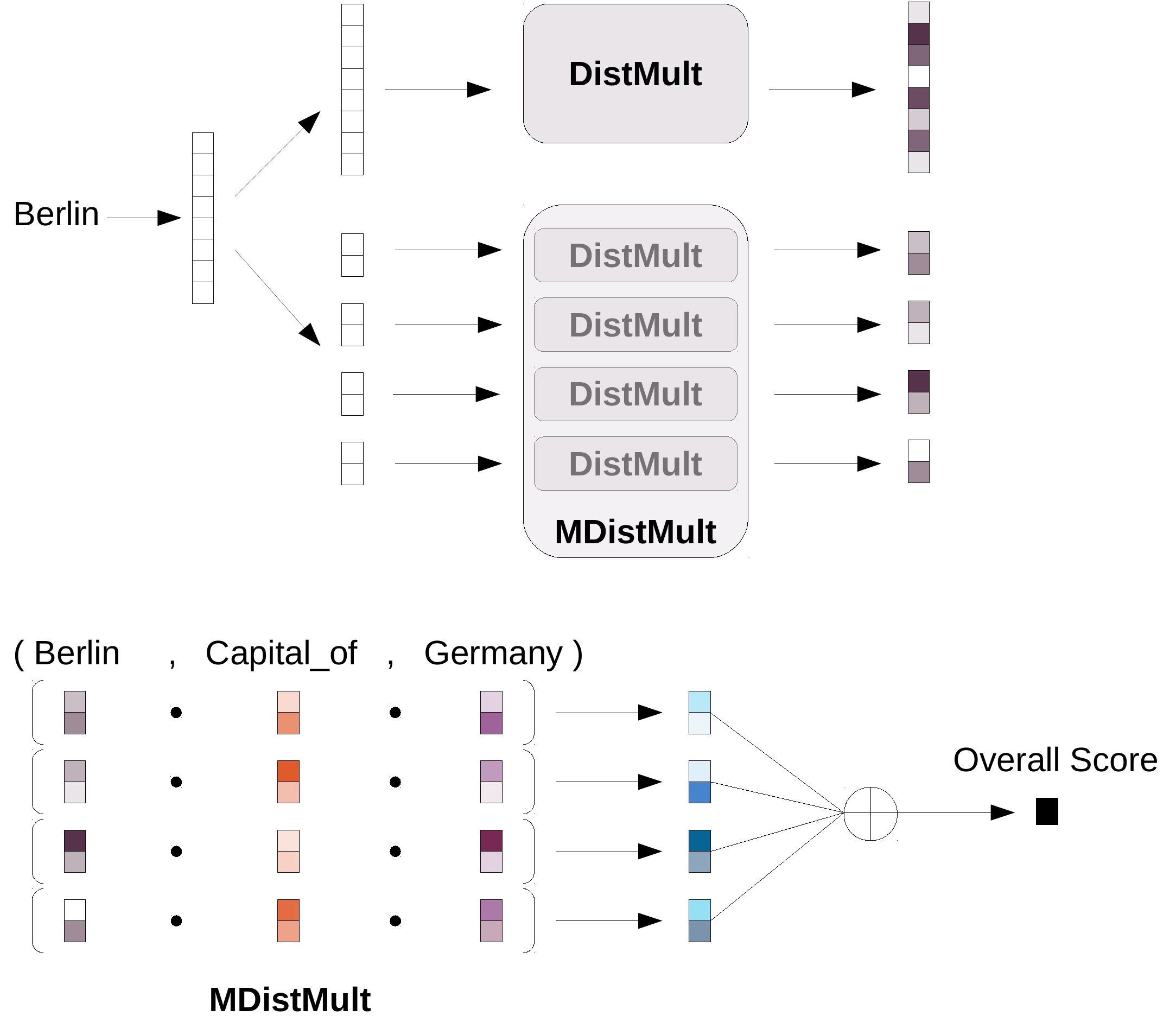} 
\caption{An example of multiple vector representation learning with MDistMult}
\label{fig:MDistMult}
\end{figure}

Figure~\ref{fig:MDistMult} gives an example of representation learning with the ensemble of multiple DistMult models, named as MDistMult. Firstly, the entity \textit{Berlin} is represented as a vector with a dimension of $k\times d_{l}$. Then the vector is divided into $k$ separate vectors with dimensions of $d_{l}$. Each separate vector is trained with a single DistMult model. Given a triple (\textit{Berlin, Capital\_of, Germany}), its overall score is equal to the average of its scores computed from the $k$ separate DistMult models.

Note that all models are trained on a same dataset with same hyperparamters and the only two difference between the models is the random initialization of the embedding vectors and the random sampling during the training process. To verify the efficiency of our approach, we keep the overall embedding size $d_l$ of multiple low-dimensional models same as the embedding size $d_h$ of the single high-dimensional model, i.e.~$k \times d_l = d_h.$ We follow the above steps to perform experiments on various state-of-the-art models, including TransE, RotatE, DistMult, ComplEx and ComplExN3. We also test DistMultN3, the extension of DistMult with the N3 regularizer. We will later elaborate on the embedding initialization and the optimization objectives of various models as well as the parallel training mechanism. 
\subsection{Initialization}
In this work, combined multiple models of the same kind are trained with same hyperparameters under the uniform experiment setting and the only two model variations are from the random initialization of embeddings before the training process and the random sampling during the training process. 

Following the previous work on KGE, we use Xavier uniform initialization for semantic matching models and Xavier norm initialization for distance-based models. Xavier~\cite{b22} initialization can make variances of each neuron layer remain the same and thus bring a faster converge speed. Xavier uniform initialization and Xavier norm initialization fill the input embedding with values drawn from the uniform distribution $\mathcal{U}(-a,a)$ and the normal distribution $\mathcal{N}(0, \sigma^2)$ respectively, where
\begin{equation}
\begin{split}
    a =  \sqrt{\frac{6}{d_{in}+d_{out}}},\quad\sigma =  \sqrt{\frac{2}{d_{in}+d_{out}}}.
\end{split}
\end{equation}
For KGE model, $d_{in}$ is equal to the embedding size $d$ and $d_{out}=0$. We follow the original experimental settings to uniformly initialize embeddings for TransE, RotatE and use Xavier norm initialization for DistMult and ComplEx. On the other hand, embeddings are initialized as sparse vectors for ComplExN3 and DistMultN3. Since the random seed is not fixed in the experiments, embeddings of different copies of the same model are initialized as different vectors.
\subsection{Optimization}
An appropriate loss function is quite important for model optimization.
Most of distance-based KGE models like TransE use a margin rank loss function as optimization objective~\cite{b6,b26,b27}. The original work introducing ComplEx obtained the state-of-the-art results with a binary logistic loss function~\cite{b14}. RotatE adds a margin parameter into the binary logistic loss function without the regularization term. This margin-based logistic loss function has been proven to be helpful to enhance the performance of distance-based models~\cite{b11, b23, b24}. In this work, we utilize the binary logistic loss function to train TransE, RotatE, DistMult and ComplEx. Given a training triple $(h,r,t)$, the binary logisitic loss function is defined as follows,
\begin{align}
   \mathcal{L}_{b} = & -\text{log}\ \sigma(\gamma+f_{\mathcal{M}}(h,r,t))-\sum_{i=1}^{\eta}p(h_{i}',r,t_{i}')\text{log}\ \sigma(-\gamma\nonumber\\
   &-f_{\mathcal{M}}(h_{i}',r,t_{i}'))+\lambda(||\mathbf{h}||_{2}^{2}+||\mathbf{r}||_{2}^{2}+||\mathbf{t}||_{2}^{2}).
\end{align}
where $f(\cdot)$ denotes the specific score function of the KGE model $\mathcal{M}$ as defined in the previous section, $\sigma(\cdot)$ denotes the sigmoid function, $\gamma$ denotes the margin of distance-based model, $\eta$ is the number of negative samples per positive ones, $(h_{i}',r,t_{i}')$ is the $i$th negative sample corresponding to $(h,r,t)$ and $p(h_{i}',r,t_{i}')$ is the weight of the negative sample which is computed from the following equation,
\begin{equation}
p(h_{i}',r,t_{i}')=\frac{\text{exp}(f_{\mathcal{M}}(h_{i}',r,t_{i}'))}{\sum_{j=1}^{\eta}\text{exp}(f_{\mathcal{M}}(h_{j}',r,t_{j}'))}.
\end{equation}
Notice that the above loss function is exactly same as the binary logistic loss function used in~\cite{b14} when $\gamma=0$ and $p(h_{i}',r,t_{i}')=1/\eta$. It is also equivalent to margin-based logistic loss function defined in~\cite{b11} when $\lambda=0$.

Different from previous work, ComplExN3 performs link prediction as a multiclass classification task by using the full negative sampling and the multi-class log-loss with N3 regularization instead of random negative sampling and the binary logistic loss with L2 regularization~\cite{b10,b25}. We follow such setting for DistMultN3. The multiclass log-loss of a training triple $(h,r,t)$ is defined as follows,
\begin{equation}
\begin{split}
        &\mathcal{L}_{m}=\mathcal{L}_{m}^{1}+\mathcal{L}_{m}^{2}+\lambda(||\mathbf{h}||_{3}^{3}+||\mathbf{r}||_{3}^{3}+||\mathbf{t}||_{3}^{3}),\\
    &\mathcal{L}_{m}^{1}=-\text{log}(\frac{\text{exp}(f_{\mathcal{M}}(h,r,t))}{\sum_{h^{'}\in\mathcal{E}}\text{exp}(f_{\mathcal{M}}(h^{'},r,t))}),\\
   &\mathcal{L}_{m}^{2}=-\text{log}(\frac{\text{exp}(f_{\mathcal{M}}(h,r,t))}{\sum_{t^{'}\in \mathcal{E}}\text{exp}(f_{\mathcal{M}}(h,r,t^{'}))}).
\end{split}
\end{equation}
\subsection{Parallel Training}
Data parallelism and task parallelism are two common forms of parallel computing. Data parallelism divides data equally among multiple processors and instantaneously execute the same function over multiple data inputs across multiple processors. 
In our case, the training process of each copy $\mathcal{M}_j$ of the model $\mathcal{M}$ is taken as a separate task. For each training task, since it is independent to each other, there is no risk of being stuck with double buffering or waiting other tasks. Thus, we can utilize task parallelism and data parallelism for training multiple KGE models at the same time in multi-GPU environments.

Multiple models can also be trained parallelly in a single-GPU environment. A single process may not utilize all the computation capacity and memory-bandwidth available on the GPU. Nvidia Multi-Process Service (MPS)~\cite{b18} allows kernel and memcopy operations from different processes to overlap on the GPU, achieving faster computation. 

\subsection{Generalization Ability}

The performance of a KGE model heavily relies on the ability of modeling various graph patterns such as symmetric, 1-n, n-1 and n-n patterns.
However, not all patterns are encoded by as single KGE model. For example, TransE as a baseline suffers from issue of modeling symmetric, 1-n, n-1 and n-n relations. 
Here we show that our approach on training KGE models can help to cope with expresivity issue of modeling patterns. 
We showcase the advantage of our approach on TransE, which is one the baselines that has been reported to suffer from various experisivity issues in modeling symmetric, 1-n, n-1 and n-n relations \cite{b19}.

\textbf{Symmetric Pattern}

Given a symmetric relation $r$ (e.g.~\textit{SimilarTo}), TransE cannot represent both $\mathbf{h} +  \mathbf{r} \approx \mathbf{t}$ and $\mathbf{t}  + \mathbf{r} \approx \mathbf{h}$ simultaneously while the relation vector is non-zero.  
Therefore, given $\mathbf{r} \neq \mathbf{0}$, TransE can express either
\begin{equation}
\begin{split}
   & \mathbf{h} + \mathbf{r} \approx \mathbf{t}, \mathbf{t} + \mathbf{r} \neq \mathbf{h}, \quad\text{or}\\
   & \mathbf{h} + \mathbf{r} \neq \mathbf{t}, \mathbf{t} + \mathbf{r} \approx \mathbf{h}.\nonumber
\end{split}
\end{equation}

Therefore, if the score of $(h,r,t)$ becomes zero, the score of $(t,r,h)$ becomes non-zero. For example, we have either case 1: $f(h,r,t) = 0, f(t,r,h) = \eta$ or case 2: $f(h,r,t) = \eta, f(t,r,h) = 0$, which none of them does not show being symmetric, even with using a very high embedding dimension (e.g.~1000). That case 1 or case 2 happens depends on the several random items used in the training (randomness in initialization, data splitting and reshuffling etc). Therefore, in different run of a single TransE, case 1 or 2 might be occurred. Therefore, a single model with any dimension cannot properly represent symmetric pattern. 
Now let us train TransE using our approach, denoted by MTransE. Assume we train two TransE models with low dimension. Because of using two different runs, in which each one has its own randomness, there is a possibility of having case 1 for the first slice of TransE (i.e.~TransE1) and case 2 for the second slice (TransE2). Therefore, we have
\begin{equation}
\begin{split}
   &  \mathbf{h}_1 + \mathbf{r}_1 \approx \mathbf{t}_1, 
    \mathbf{t}_1 + \mathbf{r}_1 \neq \mathbf{h}_1, \quad\text{and}\\
   &  \mathbf{h}_2 + \mathbf{r}_2 \neq \mathbf{t}_2,
    \mathbf{t}_2 + \mathbf{r}_2 \approx \mathbf{h}_2.\nonumber
\end{split}
\end{equation}

where $(\mathbf{h}_1, \mathbf{r}_1, \mathbf{t}_1)$ are embeddings which are associated to the first slice and $(\mathbf{h}_2, \mathbf{r}_2, \mathbf{t}_2)$ are used for the second slice. The overall score of $(h,r,t)$ computed by MTransE becomes $f(h,r,t) = 0 + \eta/2 = \eta/2$ and for $(t,r,h)$, the score becomes $f(t,r,h) = \eta/2 + 0 = \eta/2$. Consequently, it is possible for MTransE to model symmetric pattern. 

\textbf{n-1 Relation Pattern}

In addition to symmetric, TransE suffers from encoding other patterns such as 1-n, n-1 and n-n relations. Here we focus on 1-n relation encoding in TransE and our model. 

For simplicity of explanation, let n=2. 
Assume $h^1, h^2$ are two different entities satisfying $(h^1,r,t)$ and $a(h^2,r,t)$ where $r$ is an n-1 relation. Since it is less likely that $\mathbf{h}^1 = \mathbf{h}^2$. Therefore, for the relation $r$ we have either
\begin{figure}[t!]
\centering
\includegraphics[width=8.6cm]{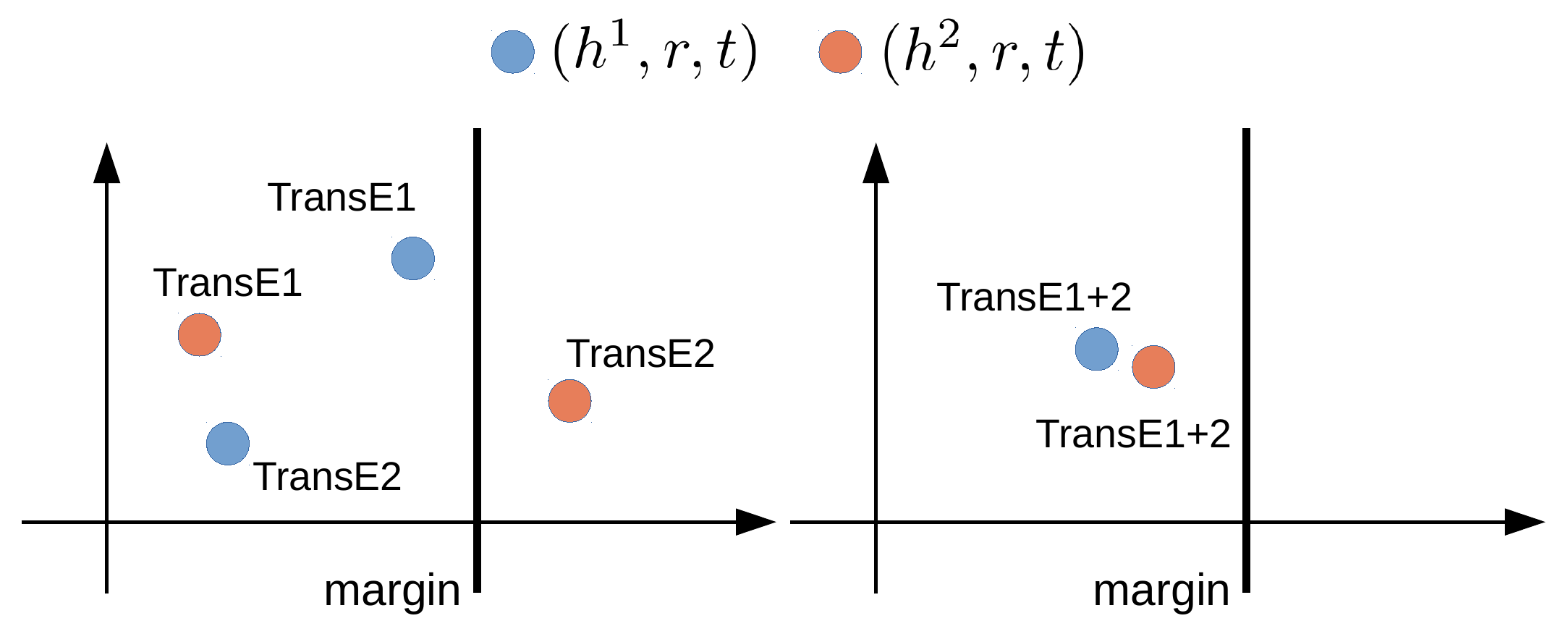} 
\caption{An example of MTransE modeling an n-1 relation}
\label{fig:MTransE}
\end{figure}
\begin{equation}
\begin{split}
  &  \mathbf{h}^1 + \mathbf{r} = \mathbf{t}, 
    \mathbf{h}^2 + \mathbf{r} \neq \mathbf{t}, \quad\text{or}\\
  &  \mathbf{h}^1 + \mathbf{r} \neq \mathbf{t}, 
    \mathbf{h}^2 + \mathbf{r} = \mathbf{t}.\nonumber
\end{split}
\end{equation}

Therefore, $(h^1, r, t)$ and $(h^2,r,t)$ get different scores while both are positive triples (ranked differently).

Using MTransE, it is possible for the model to learn 
\begin{equation}
\begin{split}
    \mathbf{h}_1^1 + \mathbf{r}_1 = \mathbf{t}_1, \mathbf{h}_1^2 + \mathbf{r}_1 \neq \mathbf{t}_1.\nonumber
\end{split}
\end{equation}
for the first slice and
\begin{equation}
\begin{split}
    \mathbf{h}_2^1 + \mathbf{r}_2 \neq \mathbf{t}_2, \mathbf{h}_2^2 + \mathbf{r}_2 = \mathbf{t}_2.\nonumber
\end{split}
\end{equation}
for the second slice. Therefore, we have $f(h^1,r,t) = 0 + \eta_{1}/2$ and $f(h^2,r,t) = \eta_{2}/2 + 0$ for ($h^1,r,t$), $(h^2,r,t)$ respectively. As shown in Figure~\ref{fig:MTransE}, these two triples have the possibility of getting more close scores and consequently are ranked more closely. The similar arguments can be used for other models and patterns.

\section{Experiments}
\subsubsection{Dataset} 
We use FB15K~\cite{b6}, FB15K-237 and WN18RR~\cite{b7} for evaluation. 
FB15K contains relation triples from Freebase, a large tuple database with structured general human knowledge. Another version of FB15K named as FB15K-237 has been created to provide a more challenging KG after removing inverse relations. WN18RR is extracted from an English lexical database, WordNet, with no inverse relations. All datasets statistics are shown in Table 1.
\begin{table}[h!]
\centering
\label{tb:datasets}
\caption{Statistics of the datasets}.
\begin{tabular}{cccccc}
\toprule
   Dataset&$|\mathcal{E}|$&$|\mathcal{R}|$&\#Train&\#Valid&\#Test\cr
  \midrule
        FB15K&14951& 1345& 483142& 50000& 59071\cr
        FB15K237& 14541& 237& 272115& 17535& 20466\cr
        WN18RR&40943 &11 &86835 &3034 &3134\cr
 \bottomrule
\end{tabular}

\end{table}

\subsubsection{Evaluation Metric}

We evaluate the performance of link prediction in the filtered setting: we first rank a test triple $(h,r,t)$ against all other candidate triples not appearing in the training, validation, or test set, where candidates are generated by corrupting the subject: $(h,r,t')$ to get the right rank of $(h,r,t)$. Likewise, the left rank of $(h,r,t)$ is its rank against candidate triples $(h',r,t)$ where $(h',r,t)\notin \Sigma_{train}\cup\Sigma_{valid}\cup\Sigma_{test}$. We use Mean Reciprocal Rank (MRR) and Hits@N for evaluation. The percentage of testing triples which are ranked lower than N is considered as Hits@N. To compute MRR, the following formula is used $\text{MRR} = \frac{1}{2n}\sum_{i=1}^{n} \frac{1}{RR_i}+ \frac{1}{LR_i}$ where $n$ is the number of testing triples, $RR_i$ and $LR_i$ are the right rank and the left rank of the $i$th test triple. To alleviate the noise from the random initialization and the random sampling, in this paper we train and evaluate each KGE model 10 times and compute the averages and the standard deviations of its MRRs and Hits@Ns, no matter whether it is a single model or a combination model. The averages of MRRs and Hits@Ns are denoted as $\overline{\text{MRR}}$ and $\overline{\text{Hits@N}}$, the standard deviations of MRRs and Hits@Ns are denoted as $\sigma_{\text{MRR}}$ and $\sigma_{\text{Hits@N}}$.

\subsubsection{Experimental Setup} 

We evaluate our proposed approach by training TransE, RotatE, DistMult, ComplEx, DistMultN3 and ComplExN3 in both single models with different dimensions and multiple models with the same dimensions. In practice, an overly small embedding size e.g., $d<100$ is not commonly used for KGE since a KGE model with such a small embedding size is typically not expressive enough to capture the semantics of entities and relations in a large-size KG, whereas a KGE with a very large dimension size suffers from over-fitting and excessive consumptions on memory size and training time. Thus, we set the embedding size to $d = \{200, 400, 600, 800, 1000, 1200\}.$ Another noteworthy point is that we define the embedding size $d$ as the total number of adjustable parameters in each embedding. For a complex-valued embedding model, e.g., ComplEx, the embedding size is as twice as the embedding dimension since each dimension has two elements of the real part and the imaginary part. To train TransE, RotatE, DistMult and ComplEx, we use an Adam optimizer with a learning rate $lr$ of 0.0003 and adopt the random negative sampling. The number of mini batches is fixed as 100, the ratio of negatives over positive training samples $\eta$ is tuned among \{$1, 3, 5, 8, 10$\}, the margin $\gamma$ of a distance-based model is tuned among $\{1,3,5,8,10,15,20,25,30\}$ and the regularization coefficient $\lambda$ of a semantic matching model is searched in $\{0.1,0.03,0.01,0.003,0.001\}$. For DistMultN3 and ComplExN3, we follow the experimental setting in~\cite{b10} to use an Adagrad optimizer with a learning rate $lr$ of 0.1 and tune the batch size $b$ in \{$100, 1000$\}, regularization coefficient $\lambda$ in $\{0.1,0.05,0.01,0.005,0.001\}$. We use the early-stop setting on validation set and set the maximum epoch to 5000.
A combination model of multiple models of a same kind is named with an initial M, e.g., an MTransE model is the ensemble of multiple TransE models.
The implementation has been done by using Pytorch on a single GPU device. \\
\section{Results and Discussion}
\begin{figure}[b]
\vspace{-0.5cm}
\centering
\includegraphics[width=8.3cm]{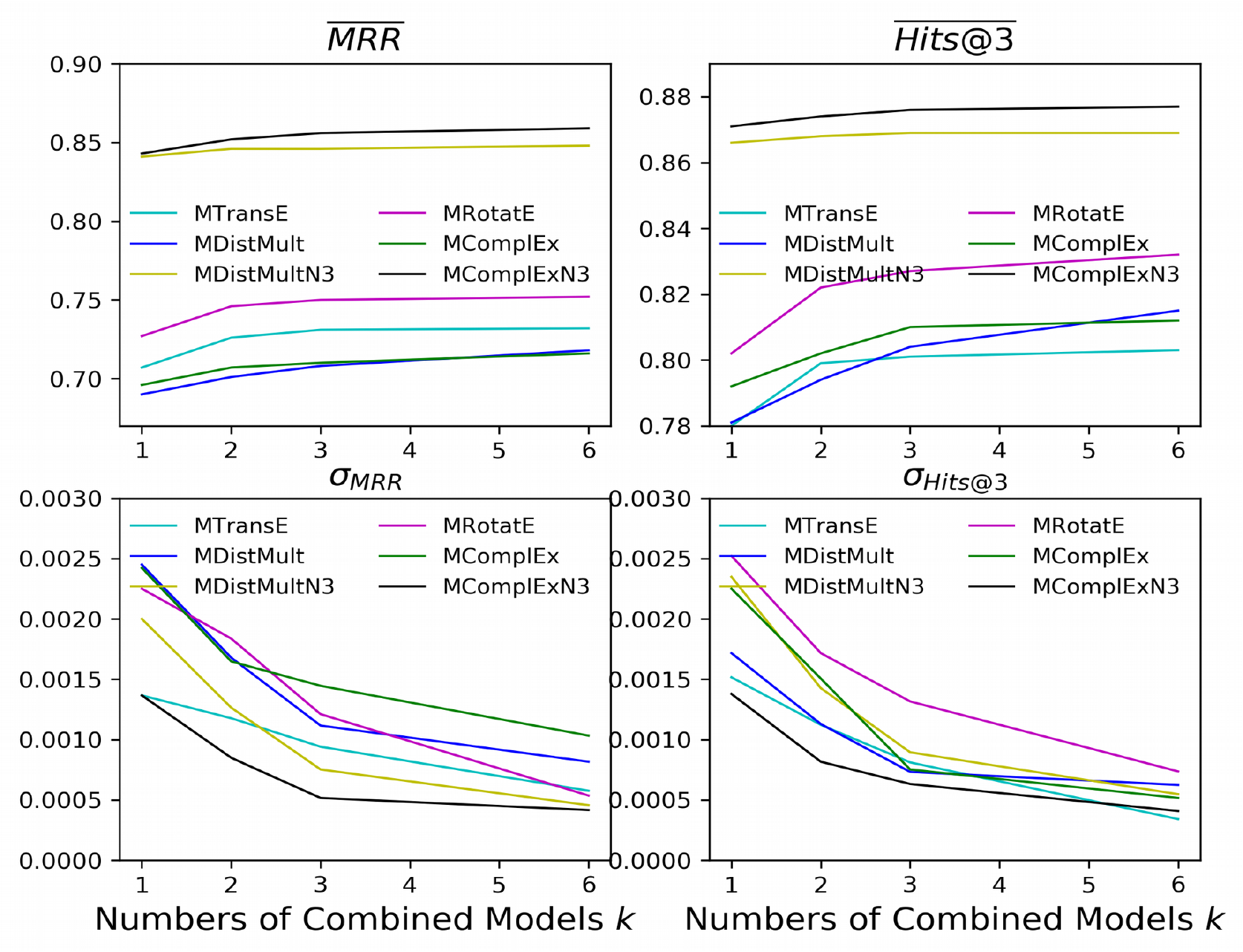} 
\caption{Link prediction results of the tested models with the same embedding size $d=1200$ and different numbers $k$ of the combined models on FB15K}
\label{fig:d1200}
\end{figure}

\subsection{Link Prediction Results}
We first study the effect of $k$ on the performance of the ensemble KGE models (e.g., MTransE, MRotatE, MDistMult, MComplEx, MDistMultN3 and MComplExN3) by fixing the overall embedding size $d=1200$ and tuning $k\in\{1,2,\dots,6\}$. As shown in Figure~\ref{fig:d1200}, when the embedding sizes are fixed as 1200, the number of combined KGE models of the same kind $k$ higher, the performance on FB15K of the combination KGE model better regarding the averages and the standard deviations of MRR and Hits@3. This observation supports the intuition that the ensemble of multiple models can alleviate the noise from the random initialization of single models.

\begin{figure*}[h]
\centering
\includegraphics[width=18cm]{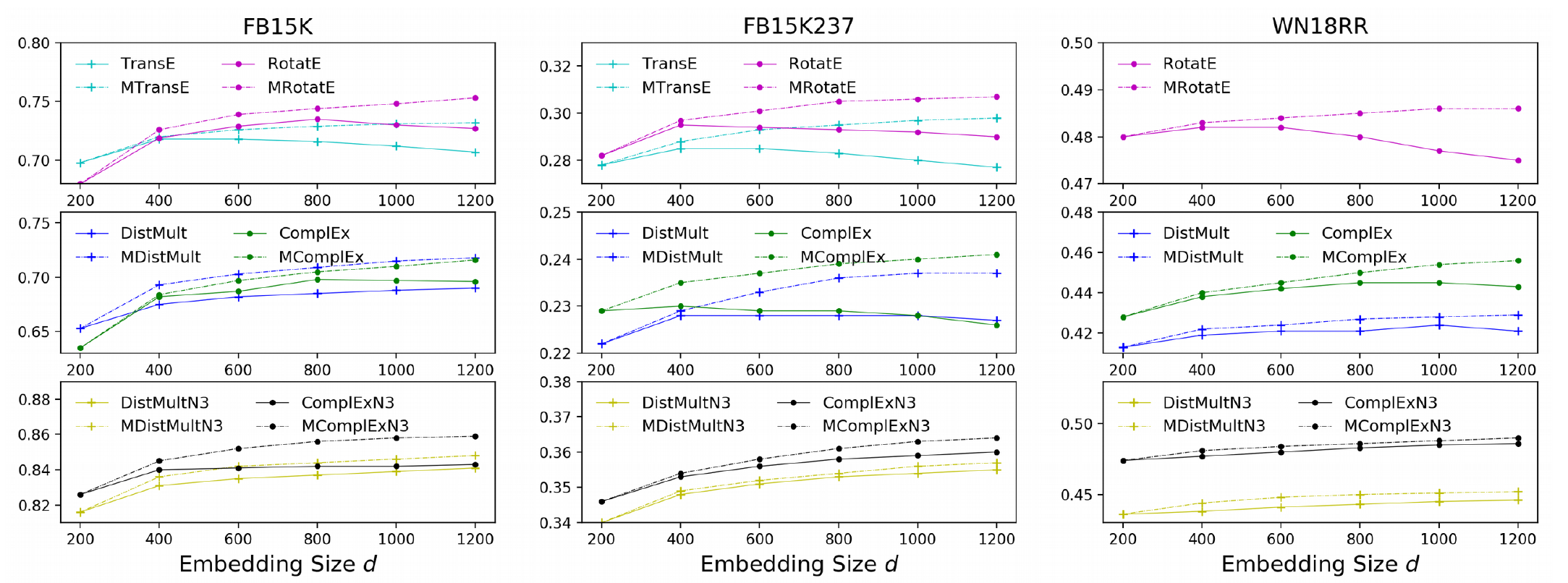} 
\vspace{-0.1cm}
\caption{$\overline{\text{MRR}}$ for link prediction of all tested models with different overall embedding sizes $d$ on KGE benchmarks}
\label{fig:MRR}
\vspace{-0.3cm}
\end{figure*}
\begin{figure*}[h]
\centering
\includegraphics[width=18cm]{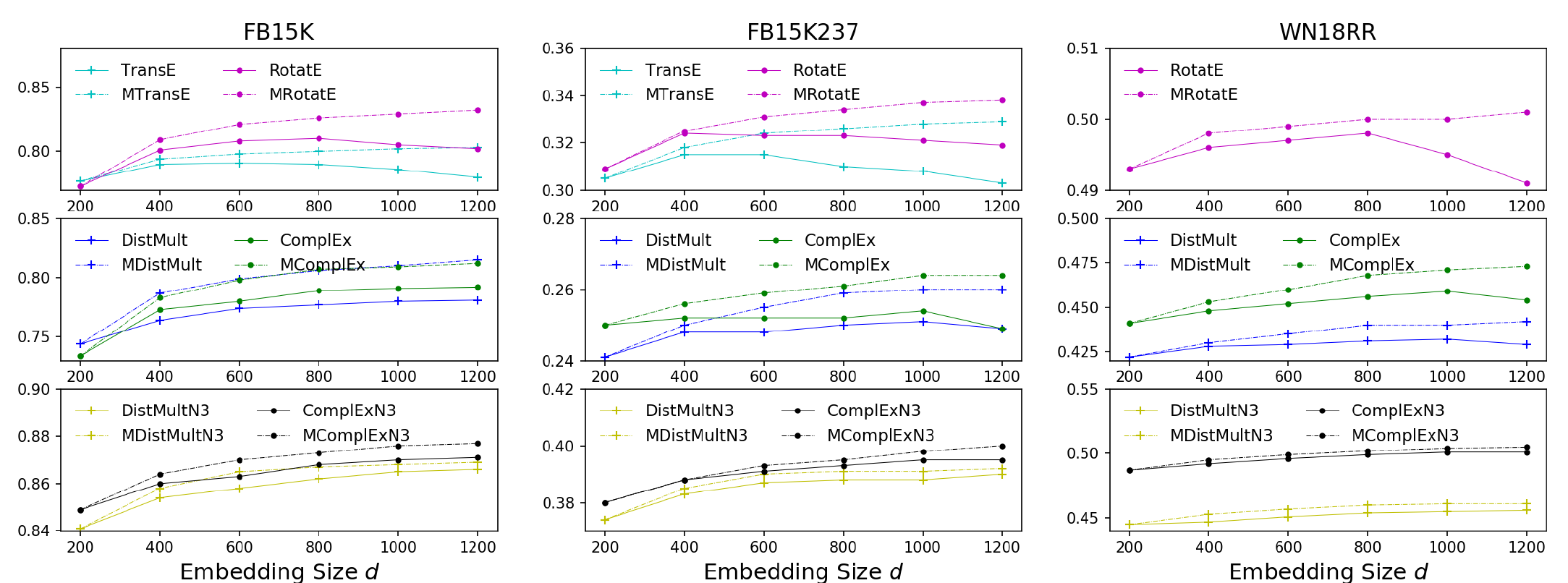} 
\vspace{-0.1cm}
\caption{$\overline{\text{Hits@3}}$ for link prediction of the tested models with different overall embedding sizes $d$ on KGE benchmarks}
\label{fig:Hits@3}
\vspace{-0.3cm}
\end{figure*}
\begin{table*}[h!]
\centering
\caption{Link prediction results on FB15K and FB15K-237.}
\begin{tabular}{lllllllll}
 \toprule
    & \multicolumn{4}{c}{FB15K}         & \multicolumn{4}{c}{FB15K-237}      \\ 
    \midrule
                          & $\overline{\text{MRR}}$ & $\overline{\text{Hits@1}}$ & $\overline{\text{Hits@3}}$ & $\overline{\text{Hits@10}}$  & $\overline{\text{MRR}}$ & $\overline{\text{Hits@1}}$ & $\overline{\text{Hits@3}}$ & $\overline{\text{Hits@10}}$ \\ \cline{2-9} 
            
                 TransE ($d=1200$) & 0.704 & 0.604 & 0.781 & 0.862 & 0.277 & 0.186 & 0.303 & 0.464 \\

                 MTransE ($d=6*200$) & \textbf{0.732} & \textbf{0.640} & \textbf{0.802} & \textbf{0.876} & \textbf{0.298} & \textbf{0.202} & \textbf{0.329} & \textbf{0.491} \\
                 \hline
                
                 DitMult ($d=1200$) & 0.688 & 0.573 & 0.781 & 
                 0.855 & 0.227 & 0.142 & 0.249 & 0.390 \\

                
                 MDitMult ($d=6*200$) & \textbf{0.718} & \textbf{0.603} & \textbf{0.815} & \textbf{0.883} & \textbf{0.237} & \textbf{0.160} & \textbf{0.260} & \textbf{0.399}  \\
                 \hline

                 ComplEx ($d=1200$) & 0.696 & 0.580 & 0.791 & 0.862 & 0.226 & 0.139 & 0.249 & 0.398 \\
            
                 MComplEx ($d=6*200$) & \textbf{0.710} & \textbf{0.590} & \textbf{0.810} & \textbf{0.886} & \textbf{0.240} & \textbf{0.162} & \textbf{0.264} & \textbf{0.400} \\
                 \hline
            
                RotatE ($d=1200$) & 0.727 & 0.630 & 0.802 & 0.868 & 0.290 & 0.197 & 0.319 & 0.478  \\
                MRotatE ($d=6*200$) & \textbf{0.753} & \textbf{0.656} & \textbf{0.832} & \textbf{0.891} &  \textbf{0.307} & \textbf{0.213} & \textbf{0.338} & \textbf{0.496} \\
                 \hline
                DitMultN3 ($d=1200$)   & 0.836 & 0.796 & 0.865 & 0.909&0.355&0.260&0.390&0.547 \\
                MDitMultN3 ($d=6*200$)   &\textbf{0.848}&\textbf{0.813}&\textbf{0.869}&\textbf{0.910}&\textbf{0.357}&\textbf{0.263}&\textbf{0.392}&\textbf{0.548} \\
                \hline
                ComplExN3 ($d=1200$)& 0.843 & 0.802 & 0.871 & 0.910&0.360&0.265&0.395&0.549 \\
                MComplExN3 ($d=6*200$)  & \textbf{0.859} & \textbf{0.829} & \textbf{0.877} & \textbf{0.911}&\textbf{0.364}&\textbf{0.268}&\textbf{0.400}&\textbf{0.555} \\
                 \hline
 \toprule 
\end{tabular}
\label{table:result_table_FB}
\vspace{-0.2cm}
\end{table*}
We further compare the link prediction performances of single high-dimensional KGE models (e.g., TransE, RotatE, DistMult, ComplEx, DistMultN3 and ComplExN3) and the simple ensembles of multiple low-dimensional KGE models of the same kinds (e.g., MTransE, MRotatE, MDistMult, MComplEx, MDistMultN3 and MComplExN3) with the same dimension sizes. The embedding size of each low-dimensional KGE models are fixed as 200 and we 
change the embedding size $d$ of an ensemble model, e.g., MTransE by tuning the number of the combined low-dimensional models $k$, i.e., $d=200\times k,k=1,2,\dots,6$.

\begin{figure*}[t]
\centering
\includegraphics[width=18cm]{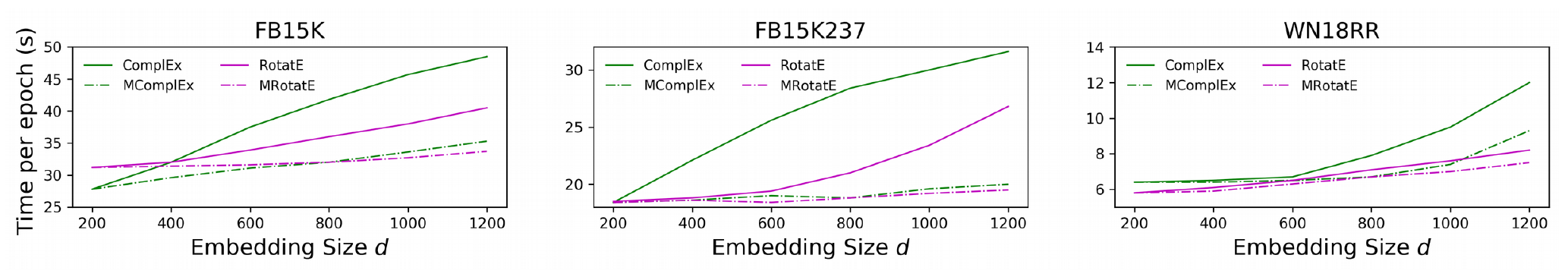} 
\vspace{-0.2cm}
\caption{Training time per epoch of (M)ComplEx and (M)RotatE with different embedding sizes $d$}
\label{fig:time}
\vspace{-0.3cm}
\end{figure*}
Figure~\ref{fig:MRR} and Figure~\ref{fig:Hits@3} show the link prediction results of the single high-dimensional models and the multiple low-dimensional models regarding $\overline{\text{MRR}}$ and $\overline{\text{Hits@3}}$. Averagely, the ensembles of multiple low-dimensional models outperform the corresponding high-dimensional single models with the same overall embedding sizes. We notice that the performance of a single KGE model might decrease with the embedding size increasing, e.g., RotatE ($d=1000$) and RotatE ($d=1200$) have lower $\overline{\text{MRR}}$s and $\overline{\text{Hits@3}}$s than RotatE ($d=800$) on FB15K. This observation demonstrates that using embeddings with an overly large embedding size leads to a risk of overfitting, especially for distance-based KGE models. By contrast, the ensemble of $k+1$ KGE models of a same kind always outperforms the ensemble of $k$ KGE models of the same kind since the new added model improves the generalization ability of the ensemble model. 

To clearly compare the ensemble KGE models and the single KGE models which have the same overall embedding sizes and scoring functions, Table~\ref{table:result_table_FB} lists the link prediction results of all target KGE models with the same overall embedding size of $d=1200$ on FB15K and FB15K237 regarding $\overline{\text{MRR}}$, $\overline{\text{Hits@1}}$, $\overline{\text{Hits@3}}$ and $\overline{\text{Hits@10}}$. For any kind of KGE approach, its high-dimensional model with $d=1200$ underperforms the ensemble of its low-dimensional models with the same overall embedding size across all metrics. For instance, MComplExN3 with $d=1200$ improves 1.6 points of $\overline{\text{MRR}}$ and 2.7 points of $\overline{\text{Hits@1}}$ compared to ComplExN3 with $d=1200$.

\subsection{Quality Analysis}
Table~\ref{tb:relation} reports the $\overline{\text{MRR}}$s of TransE ($d=200\times1$), TransE ($d=1200\times1$) and MTransE ($d=200\times6$) on link prediction involving different relation patterns in FB15K. Following the setting in~\cite{b19}, we separate relations into 4 categories: 1-1, 1-n, n-1 and n-n. Within the 1345 relations in FB15K, 24\% are 1-1, 23\% are 1-n, 29\% are n-1, and 24\% are n-n. We also employ AMIE+~\cite{b20} to extract 24 symmetric relations (r(x,y)$\Rightarrow$r(y,x)) from FB15K with a confidence threshold of 0.8. A few typical symmetric relations include \textit{person/sibling} and \textit{person/spouse}. 
As shown in Table~\ref{tb:relation}, TransE ($d=1200\times1$) have the close performance to MTransE ($d=200\times6$) on 1-1 relations, meanwhile MTransE significantly outperforms both TransE ($d=1200\times1$) and TransE ($d=200\times1$) on other complex relation patterns. MTransE can enhance the expressiveness and the generalization ability of TransE by 
comprehensively considering the link prediction results of multiple TransE models to alleviate the negative effect of TransE's limitation on modeling 1-n, n-1, n-n and symmetric relations. 
More concretely, Table~\ref{tb:examples} gives an example of how MTransE improves the performance of TransE models on modeling \textit{tv\_program/language}, an n-1 relation. Given four test triples involving the same relation \textit{tv\_program/language} and the same tail entity \textit{English} with different head entities, the first TransE model models triples (\textit{"Parks and Reactions", tv\_program/language, English}) well while the second TransE model predicts (\textit{"Angel", tv\_program/language, English}) correctly. By combining these two models, the ensemble model MTransE can predict both triples precisely. Meanwhile, the ranks of other two head entities which are underrated by both TransE models are improved by the MTransE model.
\begin{table}[t!]
\centering
\caption{$\overline{\text{MRR}}$ of (M)TransE models with different embedding size $d$ on FB15K regarding different relation patterns}.\label{tb:relation}
\resizebox{0.98\columnwidth}{!}{
\begin{tabular}{cccc}
\toprule
&TransE&TransE&MTransE\\
Relation&($d$=200$\times$1)&($d$=1200$\times$1)&($d$=200$\times$6)\\
\midrule
1-1& 0.642& 0.663($\uparrow$\textbf{0.021})&0.660($\uparrow$\textbf{0.018})\\
1-n&0.739& 0.748($\uparrow$\textbf{0.009})& 0.780($\uparrow$\textbf{0.041})\\
n-1& 0.639 &0.650($\uparrow$\textbf{0.011}) & 0.678($\uparrow$\textbf{0.039}) \\
n-n&0.706&0.709($\uparrow$\textbf{0.003}) &0.739($\uparrow$\textbf{0.033}) \\
symmetric&0.358 &0.360($\uparrow$\textbf{0.002}) &0.411($\uparrow$\textbf{0.053}) \\
 \bottomrule
\end{tabular}}
\end{table}

\begin{table}[t!]
\centering
\caption{The filtered ranks of head entities $h$ regarding the query (?, \textit{tv\_program/language, English}) obtained from TransE and MTransE models}.\label{tb:examples}
\resizebox{0.98\columnwidth}{!}{
\begin{tabular}{cccc}
\toprule
&TransE1&TransE2&MTransE\\
Head Entity ($h$)&($d$=200$\times$1)&($d$=200$\times$1)&($d$=200$\times$2)\\
\midrule
\textit{"Parks and Recreation"}& 1&5&1\\
\textit{"Angel"}&4& 1& 1\\
\textit{"Jesus of Nazareth"}& 16 &8 & 6 \\
\textit{"Nurse Jackie"}&5&6 &3 \\
 \bottomrule
\end{tabular}}
\end{table}

\subsection{Training Time}

By using MPS, we can efficiently train multiple low-dimensional KGE models on a single GPU simultaneously. We conduct all experiments on a TITAN X (Pascal) GPU. As shown in Figure~\ref{fig:time}, the time cost per epoch of training multiple low-dimensional KGE models, e.g., MRotatE ($d=200\times k$) and MComplEx ($d=200\times k$), is less than training a single high-dimensional KGE model, e.g., ComplEx and RotatE, with the same overall embedding size $d$. 



\section{Conclusion}
In this work, we empirically study the effect of the embedding size on the performance of several common KGE models and propose a performance boosting training strategy for KGE models without enlarging the overall embedding sizes of models. Concretely, we first divide a high-dimensional embedding into several low-dimensional embedding and input them into the respective KGE models of the same kind which are separately trained. All models are combined only at the query time. Given a triple ($h,r,t$), its final overall score is equal to the average of its scores computed from multiple models. We show that our approach can improve the generalization ability of KGE models on modeling various complex relation patterns. And the training processes of multiple KGE models can be completed efficiently by using parallel training. Experimental results demonstrate that the ensembles of multiple low-dimensional KGE models of the same kind outperform the corresponding single high-dimensional KGE models with the same embedding size. 


\section*{Acknowledgment}

This work is supported by the EC Horizon 2020 grant LAMBDA
(GA no. 809965), the CLEOPATRA project (GA no. 812997) and the China Scholarship Council (CSC).


\begin{thebibliography}{00}
\bibitem{b1} G. A. Miller. ``WordNet: a lexical database for English," Communications of the ACM 38, 1995, pp. 39-41.
\bibitem{b2} K. Bollacker, C. Evans, P. Paritosh, T. Sturge and J. Taylor. ``Freebase: a collaboratively created graph database for structuring human knowledge,'' In Proceedings of the 2008 ACM SIGMOD international conference on Management of data, 2008, pp. 1247-1250.
\bibitem{b3} F. M. Suchanek, G. Kasneci and G. Weikum, ``Yago: a core of semantic knowledge,'' In Proceedings of the 16th international conference on World Wide Web, 2007, pp. 697-706.
\bibitem{b4} S. Auer, C. Bizer, G. Kobilarov, J. Lehmann, R. Cyganiak, Z. Ives, ``Dbpedia: A nucleus for a web of open data,'' In The semantic web, 2007, pp. 722-735.
\bibitem{b5} Q. Wang, Z. Mao, B. Wang, L. Guo. ``Knowledge graph embedding: A survey of approaches and applications,'' IEEE Transactions on Knowledge and Data Engineering, 2017, no. 12, pp. 2724-2743.
\bibitem{b6} A. Bordes, N. Usunier, A. Garcia-Duran, J. Weston and O. Yakhnenko, ``Translating embeddings for modeling multi-relational data,'' Advances in neural information processing systems, 2013, pp. 2787-2795.
\bibitem{b7} T. Dettmers, P. Minervini, P. Stenetorp and S. Riedel, ``Convolutional 2d knowledge graph embeddings,'' In Proceedings of the AAAI Conference on Artificial Intelligence, 2018, vol. 32, no. 1.
\bibitem{b8} G. A. Miller, ``WordNet: a lexical database for English,'' Communications of the ACM, 1995, 38(11), pp. 39-41.
\bibitem{b9} K. Bollacker, C. Evans, P. Paritosh, T. Sturge and J. Taylor, ``Freebase: a collaboratively created graph database for structuring human knowledge,'' In Proceedings of the 2008 ACM SIGMOD international conference on Management of data, 2008, pp. 1247-1250.
\bibitem{b10} T. Lacroix, N. Usunier, G. Obozinski, ``Canonical tensor decomposition for knowledge base completion,'' In International Conference on Machine Learning, 2018.
\bibitem{b11} Z. Sun, Z. Deng, J. Nie, and J. Tang. ``Rotate: Knowledge graph embedding by relational rotation in590complex space,'' In International Conference on Learning Representations, 2019.
\bibitem{b23} C. Xu, M. Nayyeri, F. Alkhoury, H. Yazdi, J. Lehmann, ``Temporal Knowledge Graph Completion Based on Time Series Gaussian Embedding,'' In International Semantic Web Conference, 2020, pp. 654-671.
\bibitem{b24} C. Xu, M. Nayyeri, F. Alkhoury, H. Yazdi, J. Lehmann, "TeRo: A Time-aware Knowledge Graph Embedding via Temporal Rotation." arXiv: 2010.01029, 2020.
\bibitem{b12} S. Zhang, Y. Tay, L. Yao and Q. Liu, ``Quaternion Knowledge Graph Embedding,'' In Advances in neural information processing systems, 2019.
\bibitem{b13} B. Yang, W. T. Yih, X. He, J. Gao and L. Deng, ``Embedding entities and relations for learning and inference in knowledge bases,'' In International Conference on Learning Representations, 2014.
\bibitem{b17} M. Nickel, V. Tresp, H. P. Kriegel, ``A three-way model for collective learning on multi-relational data,'' In International Conference on Machine Learning (ICML), 2011, vol. 11, pp. 809-816.
\bibitem{b14} T. Trouillon, J. Welbl, S. Riedel, E. Gaussier and G. Bouchard, ``Complex embeddings for simple link prediction,''  In International Conference on Machine Learning (ICML), 2016.
\bibitem{b26} M. Nayyeri, C. Xu, S. Vahdati, N. Vassilyeva, E. Sallinger, H. Yazdi, J. Lehmann, "Fantastic Knowledge Graph Embeddings and How to Find the Right Space for Them." International Semantic Web Conference. Springer, Cham, 2020, pp. 438-455.
\bibitem{b27} M. Nayyeri, C. Xu, S. Vahdati, E. Sallinger, H. Yazdi, J. Lehmann, "On the knowledge graph completion using translation based embedding: the loss is as important as the score." arXiv preprint arXiv: 1909.00519 , 2019.
\bibitem{b15} D. Krompaß, V. Tresp, ``Ensemble solutions for link prediction in knowledge graph,'' In Proceedings of the 2nd Workshop on Linked Data for Knowledge Discovery, 1-10.
\bibitem{b25} C. Xu, M. Nayyeri, Y. Chen, J. Lehmann, "Knowledge graph embeddings in geometric algebras." arXiv preprint arXiv: 2010.00989, 2020.
\bibitem{b16} A. Muromägi, K. Sirts, S. Laur, ``Linear ensembles of word embedding models,'' arXiv preprint arXiv: 1704.01419, 2017.
\bibitem{b18} P. Sah, ``Improving gpu utilization with multi-process579service (MPS),''  In GPU Technology Conference, 2015, vol. 5584.
\bibitem{b19} Z. Wang, J. Zhang, J. Feng and Z. Chen, ``Know1edge Graph Embedding by Translating on Hyperplanes,'' In Proceedings of the AAAI Conference on Artificial Intelligence, 2014, pp. 1112–1119.
\bibitem{b20} L. Galárraga, C. Teflioudi, K. Hose, F. M. Suchanek, ``Fast rule mining in ontological knowledge bases with AMIE+,'' The VLDB Journal, 2015, no. 6, pp. 707-730.
\bibitem{b22} X. Glorot, Y. Bengio. ``Understanding the difficulty of training deep feedforward neural networks,'' In Proceedings of the thirteenth international conference on artificial intelligence and statistics, 2010, pp. 249-256.






\end{thebibliography}
\end{document}